\documentclass[10pt,twocolumn,letterpaper]{article}

\usepackage{cvpr}
\usepackage{times}
\usepackage{epsfig}
\usepackage{graphicx}
\usepackage{amsmath}
\usepackage{amssymb}
\usepackage{color}
\usepackage{booktabs}
\usepackage{multirow}
\usepackage{algorithm}
\usepackage[noend]{algorithmic}
\usepackage{paralist}
\usepackage{gensymb}  
\usepackage{cite}

\newcommand{\bm}[1]{\boldsymbol{#1}}

\newcommand{\be}{\mathbf{e}}

\newtheorem{lemma-ap}{Lemma}
\newtheorem{claim-ap}{Claim}

\def\be {\begin{equation}}
\def\ee {\end{equation}}
\def\beas {\begin{eqnarray*}}
\def\eeas {\end{eqnarray*}}
\def\bea {\begin{eqnarray}}
\def\eea {\end{eqnarray}}
\def\bes {\begin{equation*}}
\def\ees {\end{equation*}}

\makeatletter
\usepackage{xspace}
\def\@onedot{\ifx\@let@token.\else.\null\fi\xspace}
\DeclareRobustCommand\onedot{\futurelet\@let@token\@onedot}

\newcommand{\figref}[1]{Fig\onedot~\ref{#1}}

\newcommand{\secref}[1]{Sec\onedot~\ref{#1}}

\newcommand{\algref}[1]{Alg\onedot~\ref{#1}}

\def\eg{\emph{e.g}\onedot} 
\def\ie{\emph{i.e}\onedot} 
\def\cf{\emph{cf}\onedot} 
 
\def\wrt{w.r.t\onedot}

\makeatother

\usepackage[pagebackref=true,breaklinks=true,letterpaper=true,colorlinks,bookmarks=false]{hyperref}

\cvprfinalcopy 

\newcommand{\PAR}[1]{\vskip4pt \noindent{\bf #1~}}

\def\cvprPaperID{4167} 

\begin{document}

\title{Hybrid Scene Compression for Visual Localization}

\author{Federico Camposeco$^1$ \quad Andrea Cohen$^1$ \quad Marc Pollefeys$^{1,2}$ \quad Torsten Sattler$^3$\\
$^1$Department of Computer Science, ETH Zurich \quad $^2$ Microsoft \quad $^3$Chalmers University of Technology
}

\maketitle

\begin{abstract}
Localizing an image \wrt a 3D scene model represents a core task for many computer vision applications. An increasing number of real-world applications of visual localization on mobile devices, \eg, Augmented Reality or autonomous robots such as drones or self-driving cars, demand localization approaches to minimize storage and bandwidth requirements. Compressing the 3D models used for localization thus becomes a practical necessity. In this work, we introduce a new hybrid compression algorithm that uses a given memory limit in a more effective way. Rather than treating all 3D points equally, it represents a small set of points with full appearance information and an additional, larger set of points with compressed information. This enables our approach to obtain a more complete scene representation without increasing the memory requirements, leading to a superior performance compared to previous compression schemes. As part of our contribution, we show how to handle ambiguous matches arising from point compression during RANSAC. Besides outperforming previous  compression techniques in terms of pose accuracy under the same memory constraints, our compression scheme itself is also more efficient. Furthermore, the localization rates and accuracy obtained with our approach are comparable to state-of-the-art feature-based methods, while using a small fraction of the memory.
\end{abstract}

\section{Introduction}

Visual localization constitutes an essential step in 3D computer vision. It plays an important role in large scale Structure-from-Motion (SfM)~\cite{schoenberger2016sfm,Agarwal2009ICCV,heinly2015reconstructing} 
and SLAM~\cite{Davison07monoslam}.
Visual localization is also a key task for both robotics, \eg, self-driving cars~\cite{haene2017journal}, and mobile device  applications such as  virtual and augmented reality~\cite{Lynen-RSS-15}.

\begin{figure}[t!]
    \centering
    \includegraphics[width=0.9\columnwidth]{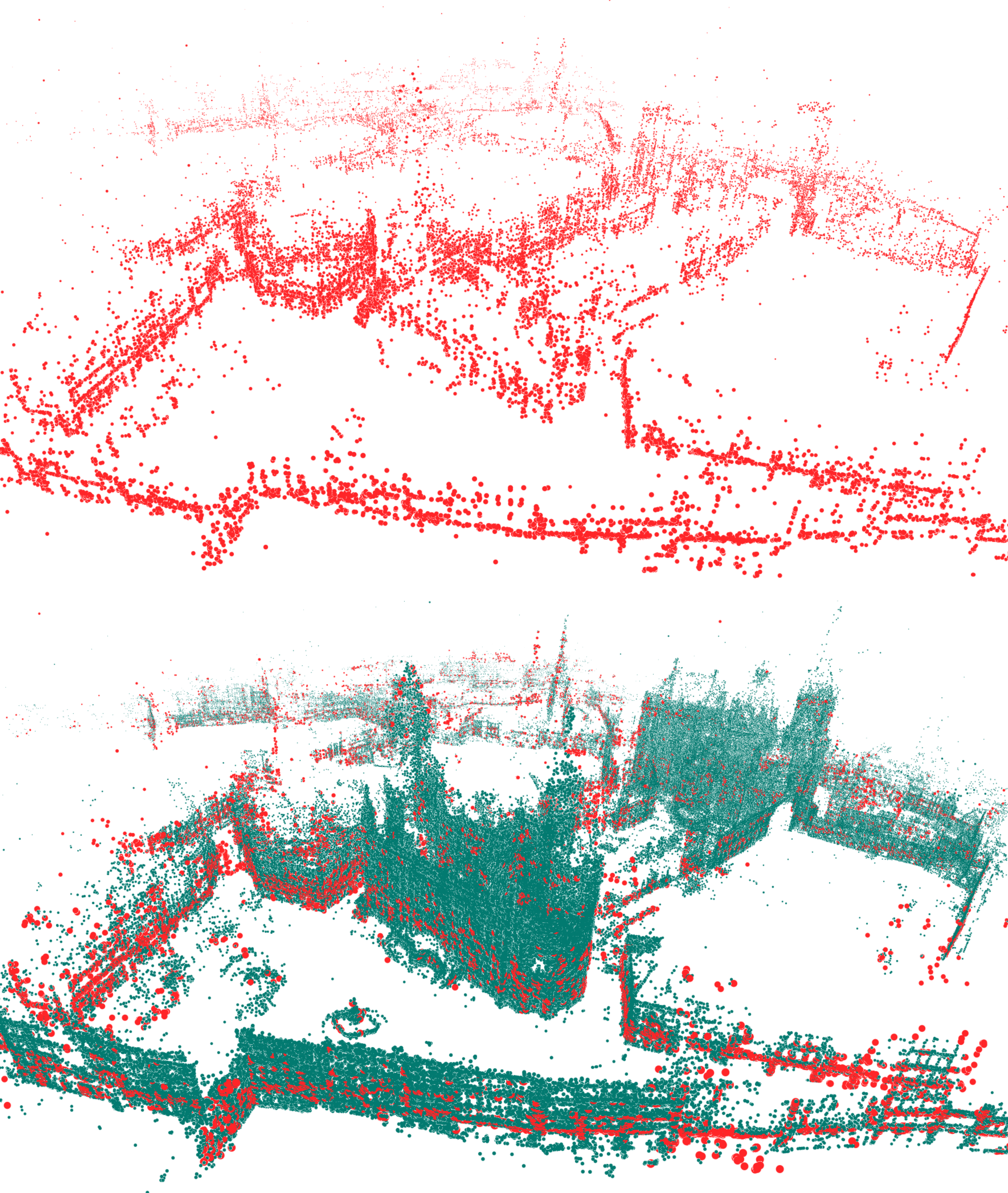}
    \vspace{1pt}
    \caption{\textbf{Comparison of compressed scenes}. (Top) 3D model obtained using a state-of-the-art model compression scheme~\cite{cao2014minimal}. (Bottom) Output of our hybrid approach. Full-descriptor points are shown in red and compressed points in green. Both models use the same amount of memory, but ours leads to a better localization performance: For a compression rate of $\sim$2\%, the percentage of localized query images improves from $\sim$70\% to more than $85$\%.}%
    \label{fig:sfms}
\end{figure}

The classical approach to visual localization requires a sparse 3D scene model, where each point is associated to a 3D position and one or more image descriptors \cite{sattler2017efficient,li2010location}.
In order to localize a query image, a set of 2D-3D correspondences can be established by using descriptor matching between features in the image and 3D points in the scene. These matches can then be used for robust pose estimation based on RANSAC and a minimal pose solver \cite{Fischler:1981:RSC:358669.358692}.

In many applications, coarse localization priors are available, \eg, from GPS or WiFi signals. 
These priors can be used to coarsely determine the part of the 3D model shown in a given query image  within tens to hundreds of meters. 
Thus, it is often only necessary to match against a part of a larger 3D model. 
Still, it is important to compress these small- to medium-scale parts, \eg, to be able to transfer them to mobile devices or to be able to handle more requests in a server-side application~\cite{Middelberg14ECCV}. In this work, we thus focus on efficiently handling scenes of this size.

Previous works on 3D scene compression mostly fall under two categories: either the models are compressed by selecting a subset of all original 3D points while keeping their full visual information (\ie, full feature descriptors) \cite{cao2014minimal,li2010location,marcin2015ICRA}, or all or a subset of the points are kept but their descriptors are compressed~\cite{Sattler_2015_ICCV,Lynen-RSS-15}. On a different line of work, CNN-based localization approaches such as PoseNet \cite{kendall2015posenet,walch2017image,kendall2016modelling,kendall2017geometric} or DSAC~\cite{Brachmann2017CVPR,Brachmann2018CVPR} implicitly represent a scene by the weights stored in a network and can thus also serve as compression methods. However, we show that our approach outperforms CNN-based methods in terms of pose accuracy, memory consumption, or both. 
Our approach is also more flexible in the presence of scene changes~\cite{sattler2017large} as no expensive retraining of a CNN is required.

In this paper, we present a hybrid 3D scene compression method that selects two subsets of scene points: the first subset consists of very few, carefully selected points with full descriptors, while the second, larger subset consists of points associated to quantized descriptors. The first set of points is used to obtain high-quality correspondences via full descriptor matching that can be used to generate pose hypotheses inside RANSAC. As the number of points in this set is very small due to memory requirements, they might not lead to enough matches for pose verification. Therefore, the second (and larger) set of points for which we only store compressed descriptors is used. While the resulting matches are not unique and are thus not well suited to generate pose hypotheses, they are sufficient for pose verification. Due to compressing the descriptors of the second set, our approach allows us to select significantly more points at the same memory consumption as standard approaches that store full descriptors per 3D point. One central insight of this paper is that selecting more points for pose verification is important for accurate pose estimation. We thus show that storing these two sets of points provides more geometric information for localization while keeping the same memory consumption as other compression methods. An example of our compression output is  visualized on the right in \figref{fig:sfms}.

This work also introduces a RANSAC variation for robust pose estimation that exploits the two types of 3D point subsets, as well as 3D point co-visibility by using a guided sampling strategy (inspired by \cite{sattler2017efficient,li:eccv:2012}) that aims to choose only co-visible 3D points to produce minimal subsets.
This sampling strategy increases the chances of finding a good minimal sample, which can be critical for accuracy and performance when dealing with lower inlier ratios.

In short, this paper presents the following contributions:
    \textbf{1)} We propose a novel hybrid 3D scene compression algorithm that takes into account the visibility of the 3D points, their coverage of the scene and the visual uniqueness of their appearance. Both coverage and uniqueness are enforced using a novel technique that allows for better coverage while offering faster compression times than the state-of-the-art.
    \textbf{2)} We obtain two different types of compressed points: spatially compressed but fully descriptive points that result in good unique matches, and a larger set of points compressed in appearance space that result in multi-matches. This second set of points has a low memory consumption while being very helpful for avoiding a loss in localization performance. 
    \textbf{3)} We introduce a novel RANSAC variant that exploits multi-matches for model evaluation during robust pose estimation and uses a co-visibility-based sampling prior. This modified RANSAC improves both the localization rate and the pose accuracy.
    \textbf{4)} We validate our method on six small- to medium-scale datasets, showing that our compression scheme outperforms state-of-the-art compression methods both in terms of number of localized images, as well as pose estimation accuracy. Our approach is competitive with state-of-the-art localization approaches while using significantly less memory. 

\section{Related Work}
\label{sec:rel_work}
The literature on 3D scene compression for visual localization can be divided into three main categories: image retrieval, neural network-based scene representations, and scene compression for feature-based localization methods.

Image retrieval techniques \cite{Sattler_2015_ICCV,zhang2006image,philbin2007object} perform compression by representing each image through a set of visual words (or using an even more compact representation such as the VLAD descriptor \cite{Arandjelovic:2013:VLA:2514950.2516209,Torii2015CVPR,Arandjelovic16}), meaning that only little data needs to be kept in memory at all times. These methods may provide an approximate pose estimate via the known pose of the retrieved database images. In order to improve retrieval performance, \cite{Irschara09fromstructure-from-motion} synthesizes a set of minimal views from the 3D scene in order to cover the whole scene and reduce memory requirements. As a next step, the data needed for accurate pose estimation can either be loaded from disk on demand (\eg 3D points and their descriptors)~\cite{Cao13CVPR,Irschara09fromstructure-from-motion}, or the local structure can be recomputed on the fly \cite{sattler2017large}. As such, the retrieval step is typically efficient, but the next stages are either slow due to loading from disk (which can be alleviated by using compressed models) or due to heavy computations (solving a local structure-from-motion problem). Especially the latter methods trade in memory compression rate for run-time and are not fit for embedded applications.

Rather than matching local features, approaches for pose~\cite{kendall2017geometric,kendall2015posenet,kendall2016modelling,walch2017image,Naseer2017IROS} or scene coordinate regression~\cite{Brachmann2017CVPR,Brachmann2018CVPR,Massiceti17CVPR} use CNNs for localization. 
They thus store information about the 3D scene implicitly in the network rather than explicitly in a point cloud. 
Pose regression approaches such as PoseNet and its variants~\cite{kendall2017geometric,kendall2015posenet,kendall2016modelling,walch2017image} claim to offer compact scene representations. However, our experiments show that our approach outperforms this type of methods both in terms of memory consumption and pose accuracy. 
Scene coordinate regression approaches predict the coordinates of a corresponding 3D point for a given patch~\cite{Brachmann2017CVPR,Brachmann2018CVPR,Massiceti17CVPR,Shotton2013CVPR}. 
They thus replace the descriptor matching stage of classical approaches through machine learning. 
While they set the state-of-the-art in pose accuracy in small-scale scenes~\cite{Brachmann2018CVPR}, they require significantly more memory than our approach.

The approach presented in this paper falls under the third category: scene compression for localization methods based on 2D-to-3D descriptor matching. Such methods compress scenes by selecting a subset of points \cite{soo20133d,li2010location,cao2014minimal,marcin2015ICRA} or by compressing the point descriptors~\cite{Sattler_2015_ICCV,Lynen-RSS-15}. Our method introduces a hybrid approach, showing that we can combine a set of points with full descriptors with a larger set of points with quantized descriptors. \cite{Lynen-RSS-15} presents an approach which combines both point selection with descriptor quantization. They use both compression techniques on top of each other, maintaining only one small set of points with quantized descriptors. This double compression drastically reduces the localization rate. They compensate by camera pose tracking over multiple frames. In contrast, we are interested in compressing the scene with minimal loss in localization performance when using a single query image.

\cite{li2010location} formulates scene compression as a $K$-covering problem: 
The newly compressed scene is composed of a subset of  points with high visibility, selected such that at least $K$ of them are seen from each of the database images used to construct the scene. 
The visibility of a 3D point is defined as the number of database images that participated in the reconstruction of this point.
A point with high visibility usually has a more accurate position and a higher probability of being observable from a large set of viewpoints~\cite{li2010location}. 
Ensuring that at least $K$ points are seen in each database image ensures that the compressed scene  covers the same area as the original scene. 
Our work modifies the $K$-covering formulation in order to enforce a more uniform distribution of the selected 3D points when projected into the database images. Instead of trying to cover $K$ points per image, we first divide each image into $q$ uniformly-sized cells and require the compressed scene to cover $K/q$ 3D points per cell. This improves localization performance \wrt previous methods for the same compression rates without an increase in compression times. 

Cao and Snavely \cite{cao2014minimal} show that, although visibility is important when choosing the best subset of points to represent the 3D scene, the  distinctiveness, or visual uniqueness, of the point descriptors should also be taken into account during compression. 
This ensures that points with an unique descriptor are selected, which improves matching performance, especially for high compression rates.  Distinctiveness is introduced to the compression by checking that each new point added to the compressed scene has a minimum descriptor distance to all the points that have already been chosen. This procedure of comparing each new point to all previously selected points is computationally  expensive. 
In this work, we also extend the $K$-covering algorithm by taking distinctiveness into account. We show that it is sufficient to approximate similarity via quantization rather than explicitly computing descriptor distances. We exploit quantization both during $K$-cover to choose a proper set of distinctive and  descriptive 3D points, as well as for selection and appearance compression of the second, larger subset of 3D points. The use of quantization allows us to decrease compression run-times considerably \wrt \cite{cao2014minimal}. 
As such, our approach is well-suited for scenarios in which the scene model needs to be re-compressed frequently, \eg, in dynamic scenes in which the geometry of the scene changes over time. 
At the same time, our hybrid sets of points increase localization performance \wrt \cite{cao2014minimal}.

Similarly to our method, \cite{Sattler_2015_ICCV} also exploits one-to-many 2D-3D matches or multi-matches via quantization. Yet, they attempt to resolve these ambiguities before pose estimation by ensuring that matches are locally unique. 
Non-unique matches are simply discarded. 
In contrast, our work actively uses ambiguous multi-matches during the hypothesis evaluation step of pose estimation. 
This eliminates the chance of rejecting correct matches before RANSAC. \cite{Sattler_2015_ICCV} requires a large codebook of 16M words to avoid introducing too many ambiguous matches. 
Thus,~\cite{Sattler_2015_ICCV} can be used to compress large-scale scenes, but is unsuitable for the smaller scene fragments used in practice~\cite{sattler2017large}. 

\cite{McIlroy2010BMVC} propose a RANSAC variant that handles multi-matches based on matching probabilities (computed from matching scores) and their involvement in (un)successful  hypotheses. 
Their method is not directly applicable in our setting due to our binary similarity measure based on whether descriptors fall into the same visual word. 
But sampling multi-matches during hypothesis generation in our RANSAC if they were inliers to previously generated poses  
could potentially allow even higher compression rates.

\section{3D Scene Compression}
\label{section:compression}
A 3D scene is composed of 3D points and database images. Each 3D point is associated to a set of SIFT descriptors corresponding to the image features from which the point was triangulated. These descriptors can then be averaged into a single SIFT descriptor that describes the appearance of that point to reduce memory consumption~\cite{li2010location}.

In order to localize a given image \wrt the 3D scene, 2D-to-3D matches are first established, which are then used in RANSAC-based pose estimation~\cite{Fischler:1981:RSC:358669.358692}.
As in \cite{sattler2017efficient}, we employ a vocabulary-based approach to feature matching. Using a pre-built vocabulary-tree, we first assign each 3D average descriptor to its corresponding K-means cell (visual word).
At search time, each query image descriptor is assigned to its closest visual word $w$. 
We then select every 3D point in the scene that has the same visual word and search for a nearest neighbor in descriptor space among those selected points.
As it will be seen next, we will  make use of this vocabulary tree again during compression.
This will allow us to both compress and query the 3D scene with a single visual vocabulary. 
This  
yields faster compression times and better localization performance \wrt previous work. 

After averaging the SIFT descriptors, most of the memory consumption of the 3D scene is concentrated in the averaged descriptors of the points.
Therefore, we will aim to reduce memory consumption by \begin{inparaenum}[1)]\item reducing the set of 3D points and \item compressing the descriptor of a subset of non-previously selected points using a visual word.\end{inparaenum}

\subsection{Reducing the Number of 3D Points}
\label{subsection:reducing}
Let the initial set of 3D points be $\mathcal{P}$.
The goal of compression is to select a subset $\mathcal{P}' \subset \mathcal{P}$ such that $|\mathcal{P}'|$ is minimal under the condition that $\mathcal{P}'$ can be used to localize as many query images as possible.
To tackle this problem, we begin with the assumption that the spatial distribution of query images will be close to the distribution of the database.
This common assumption~\cite{li2010location,Camposeco2017CVPR} is sensible given that local features are not invariant to viewpoint changes.

We aim to select points such that each of the database cameras sees \emph{enough} 3D points, \ie, we want to enforce that each of the cameras in the database are \emph{covered} by at least $K$ points. This problem of choosing the optimal set of 3D points that cover all database cameras is an instance of the Set Cover Problem (or $K$-cover when at least $K$ points must be seen from each camera)~\cite{li2010location}, which is NP-hard.
Furthermore, 
by choosing two 3D points whose descriptors are too similar, a query image descriptor could be matched to either of those points, resulting in ambiguities.
This problem arises in the presence of repeated structures that produce very similar descriptors that are actually meters apart. Therefore, the \textit{visual uniqueness} of the selected 3D points should also be taken into account.

Our primary concern is to select points which have a complete coverage of the scene, while penalizing points whose descriptors can be confused with other descriptors selected.
We can thus, as proposed by \cite{cao2014minimal}, cast this problem into an instance of the Weighted $K$-cover problem \cite{lim2014lazy}, where  
the weight reflects the discriminative power of the descriptor associated with a point.
Since the Weighted $K$-cover problem is NP-hard, we will follow common practice~\cite{li2010location,Irschara09fromstructure-from-motion,sattler2017efficient} and employ a greedy approach in order to arrive at an approximate solution.

For compression, we make use of the visibility graph~\cite{li2010location} defined by each SfM model. 
The nodes of this bipartite graph correspond to the 3D points and database images in the reconstruction. 
The graph contains an undirected edge between a point node and an image node if the point was triangulated from the image. 
Let the binary matrix $M$ represent the visibility graph 
of the SfM model\footnote{In practice, $M$ is very sparse and typically stored as an adjacency list.}, where $M_{i,j}$ is $1$ if the $i$-th image $I_i$ observes the $j$-th point $p_j$.
Starting from an empty set of points $\mathcal{P}'$, our objective at each iteration is to find $p_j$ that maximizes the gain 
\begin{equation}
\mathcal{G}(p_j, \mathcal{P}') = \bm{\alpha}(p_j,\mathcal{P}')\cdot\sum_{I_i \in \mathcal{I} \setminus C}M_{i,j} \enspace , 
\end{equation}
where $C$ is the set of images that have already been covered by $K$ points. 
The weights $\bm{\alpha}(p_j,\mathcal{P}')$ 
measure 
how visually distinctive point $p_j$ is \wrt the already selected points $\mathcal{P}'$.

In \cite{cao2014minimal}, $\bm{\alpha}$ is computed by comparing the descriptor of each candidate $p_j$ against all of the already selected 3D points.
This produces a rather slow procedure, since each descriptor comparison can be costly and an adaptive search structure (such as a KD-Tree) cannot be used as the size of $\mathcal{P}'$ increases with each iteration.

Instead, we exploit the fact that we perform vocabulary-based image localization using a pre-built vocabulary tree, \ie, 
we assign each 3D point to a visual word before 2D-to-3D matching.
Thus, we define the weighting term as 
\begin{equation}
\bm{\alpha}(p_j,\mathcal{P}') = 1 - \frac{|P_w'|}{\beta} \enspace .
\end{equation}
Here, $w$ is the visual word of the descriptor of point $p_j$, $P_w'$ is the set of selected 3D points with descriptors assigned to $w$, and $\beta$ is the maximum number of allowed points per word (set to 10 in our experiments).
We thus penalize the inclusion of 3D points whose visual word is too populated with a linear function. The intuition behind this is that the number of points assigned to a visual word gives an idea of how unique points in this visual word are.

In contrast to \cite{cao2014minimal}, we opt to enforce that not only each image is covered by $K$ 3D points but that the distribution of those points on the images is as even as possible. 
We thus subdivide each database image into a grid with $q$ equal-area cells.
We then regard the image cells instead of the images themselves as the elements to be covered by the 3D points. 
Let $c_h\in\mathcal{I}^c$, where $h=1,\dots, q\times N$ and $N$ is the total number of database images, be the $h$-th image cell to be covered. $\mathcal{I}^c$ represents the set of all images cells and $C^c$ the set of cells already covered. The gain to maximize in each iteration thus becomes 
\begin{equation}
\mathcal{G}(p_j, \mathcal{P}') = \bm{\alpha}(p_j,\mathcal{P}')\cdot\sum_{c_h \in \mathcal{I}^c \setminus C^c}M_{i,j}\enspace . \label{eq:image_space}
\end{equation}
Although the number of ``cameras'' to cover increases to $q\times K$, the number of non-zero entries remains the same. 
We thus found no noticeable increase in compression run-time since we also reduce the number of 3D points that are enforced to be viewed by each image cell to $K/q$.
As we will show in Sec. \ref{sec:experiments}, our approach has a positive impact on the localization rates: the point selection has a more even coverage of the scene and prevents the selection process to be biased towards highly structured or textured parts of the scene, which might not necessarily be visible at query time. Additionally, a uniform distribution in 2D typically leads to more stable pose estimates compared to finding all matches in a single region of an image~\cite{Irschara09fromstructure-from-motion}.

Our approach achieves the same complexity reduction as previous work: for $n$ points and $m$ images, the average number of points per image is $n/m$. 
$K$-cover methods such as~\cite{li2010location,cao2014minimal} reduce this number to $\Theta(K)$. 
The space reduction for any constant $K$ is thus $\Theta(n / (K\cdot m))$, \ie, linear in the number of 3D points.  
Eq.~\ref{eq:image_space} replaces $m$ images by $q\cdot m$ cells and $K$ with $K/q$, resulting in the same space complexity.

Once the greedy Weighted Set Cover algorithm is completed, we are left with a subset of 3D points $\mathcal{P'} \subset \mathcal{P}$ which should maximize scene coverage and descriptor uniqueness.
Instead of discarding the rest of the 3D points~\cite{cao2014minimal}, we choose to select a second subset $\mathcal{P''} \subset \mathcal{P}$ of points for which only a compressed descriptor will be kept. This procedure is detailed next.

\subsection{Selecting Multi-Matches}
\label{subsection:selecting}
As previously mentioned, the most memory-consuming part of a 3D scene are the feature descriptors.
As such, we may only select a small number of them to ensure that we substantially reduce the amount of memory required to represent a 3D model.
If we only keep the 3D points selected by our Set Covering procedure, we are prone to miss matches due to the imperfect nature of the descriptors (features might not match due to large viewpoint changes) and the matching procedure. Additionally, only a few points might be visible in the query image as the database images are only an approximation to the set of all viewpoints.
To mitigate this, we select a second subset of 3D points $\mathcal{P}'' \subset \mathcal{P}\setminus\mathcal{P}'$, where we will keep only the quantized descriptor (word assignment) for each selected 3D point and its 3D location. The purpose of this second subset of \emph{compressed points} is the following: while the points in $\mathcal{P}'$ result in high-quality matches that can be used for generating pose hypotheses during RANSAC, they might not be enough to properly verify these hypotheses. The set $\mathcal{P}''$ is thus used for verification. As these last matches are only used with a given hypothesis, they do not need to be unique as they are disambiguated by the pose. Therefore, they can be stored at little memory consumption.

We note that for each full 3D point we must store the 3D point position, the full mean descriptor, and a list of cameras that see the 3D point (which will be used within RANSAC to discard bad samples, see Sec.~\ref{section:ransac}).
For compressed points, we only need to store the 3D position and one visual word index (1 integer).
For 128 byte SIFT descriptor and a 32-bit visual word, we measure that on average we can store 26.5 compressed descriptor for each full descriptor.
Since the only way of distinguishing compressed points is by their visual word, their selection process will thus focus on word uniqueness.
To select the points for compression, we use the following procedure: \begin{inparaenum}[1)]\item score each point by its word occupancy (assigning a high score to low occupancy) and \item select the $X$ points that have the highest score. \end{inparaenum}
This procedure selects points without any regard to their camera coverage. 
However, this criterion has already been prioritized by the points selected in the first step of the compression.

Since we will, in general, have more than one 3D point with the same visual word, we must consider each 2D-to-3D match against compressed points as a \emph{multi-match} (which will be handled differently within RANSAC, as explained in the next section).
Nevertheless, the selection procedure for our compressed 3D points already minimizes the number of points in each word. 
Thus, we can expect to have a low number of multi-matches per query descriptor.

\section{Multi-Match RANSAC}
\label{section:ransac}
As explained in \secref{section:compression}, matching the 2D features of the query image against $\mathcal{P'}$ using a standard ratio test \cite{lowe2004distinctive} provides a set of one-to-one 2D-3D matches. 
We will denote this set of matches as $\mathcal{U}$. 
Matching against $\mathcal{P''}$ results in a set of multi-matches, denoted as $\mathcal{W}$. The total set of matches will be referred to as $\mathcal{M}=\{\mathcal{U}\cup \mathcal{W}\}$.
This section explains how $\mathcal{M}$ can be used during the robust RANSAC-based pose estimation stage in order to improve both localization rates and run-time when dealing with highly compressed scenes. \algref{alg:ransac} summarizes our approach. 

Robust pose estimation is usually achieved by running RANSAC \cite{Fischler:1981:RSC:358669.358692} with a minimal geometric pose solver (\eg P3P \cite{Kneip:2011:NPP:2191740.2191939} for the calibrated setting or P4P \cite{BujnakKP08} for the unknown focal length case). A general RANSAC approach randomly selects minimal samples from $\mathcal{M}$, generates pose hypotheses with the minimal solver, and then evaluates these poses by counting the number of inliers according to a given threshold on the reprojection error. However, the lower the inlier ratio, the more samples are required by RANSAC to ensure that the correct pose is found with a certain probability. Indeed, the number of samples required increases exponentially with the outlier ratio. 
Thus, using the multi-matches $\mathcal{W}$ during the sampling stage potentially introduces a high number of outliers, resulting in a prohibitively large number of RANSAC iterations. 
Therefore, our RANSAC variant limits the sampling only to the set $\mathcal{U}$ of unique matches, since these matches have a higher probability of being actually good matches. Once a model has been sampled using only unique matches, it needs to be evaluated. Having a larger number of inliers increases the confidence in the quality of a sampled model. Therefore, the whole set of matches $\mathcal{M}$, including the multi-matches $\mathcal{W}$, is used to evaluate each  pose. 

In order to further reduce the number of RANSAC iterations, previous methods such as \cite{chum2005matching} use a guided-sampling strategy in order to increase the chances of finding a good sample. Inspired by this strategy, we propose an efficient variation of the method introduced by \cite{li:eccv:2012} in the form of a simple co-visibility-based sampling strategy. This addresses the problem that even in $\mathcal{U}$ inlier ratios can be very small in a highly compressed scene and that it is therefore not enough to just limit sampling to this subset. It is desirable that samples should be drawn from sets of matches that correspond to co-visible points, \ie, those that are seen from the same camera or neighboring cameras, as these matches have a higher chance of being geometrically consistent. Similarly to \cite{li:eccv:2012}, our sampling should thus increase the probability of drawing from such subsets of matches.

More specifically, given $\mathcal{U}=\{s_j\leftrightarrow	p_j\}$, where $s_j$ denotes a 2D feature and $p_j$ a 3D point, and $n$ the cardinality of a minimal sample $\mathcal{S}\subset \mathcal{U}$ drawn in RANSAC, we would like all points $p_i \in \mathcal{S}$ to be co-visible. In \cite{li:eccv:2012}, points are considered to be co-visible if they are all observed together in at least one database image. However, we found this definition of co-visibility to be a rather slow at sampling time due to the need to perform multiple set intersection, requiring about 20ms per RANSAC iteration.

We therefore use a slightly different definition of co-visibility: for a set of points $\mathcal{Q}=\{p_i, i=1\dots n\}$, let $\mathcal{C}(p_i), p_i\in\mathcal{Q}$, be the set of database images in the scene from which point $p_i$ was reconstructed. 
The set $\mathcal{Q}$ can be represented as a graph $\mathcal{G_Q}$, where each point $p_i$ is a node and an edge is added between a pair of points $p_i, p_j, i\neq j$, if they share at least one image in the scene, \ie, if $\mathcal{C}(p_i) \cap \mathcal{C}(p_j) \neq \emptyset$. We consider the points in $\mathcal{Q}$ as co-visible if each pair of points in $\mathcal{Q}$ are at most $2$ edges apart in the graph $\mathcal{G_Q}$. 

This definition allows the efficient sampling ($\sim$1$\mu$s) of sets of covisible points: 
we choose a random match $m_1 = s_{j_1} \leftrightarrow p_{j_1}$ from $\mathcal{U}$ as the first element of $\mathcal{S}$. We sequentially draw the following $m_i$, $i=2\dots n$, samples and test whether they have an image in common with $p_{j_1}$. If they do, then the sample $m_i$ is accepted and included in $\mathcal{S}$. Otherwise, $m_i$ is dropped and re-sampled. 
If no valid $m_i$ can be found after $F$ attempts, we drop the complete sample $\mathcal{S}$. Both for speed and simplicity, we choose to always look for intersections with $\mathcal{C}(p_{j_1})$ instead of checking all of the points that are already part of $\mathcal{S}$. 
Since the first match is drawn uniformly at random, this choice does not restrict sampling. 
As a result of our simple sampling strategy, only potentially good minimal subsets are used for pose estimation and evaluation.

\begin{algorithm}[t]
\scriptsize
\algsetup{linenosize=\scriptsize}
\caption{Modified RANSAC}
\label{alg:ransac}
\begin{algorithmic}[1]
\REQUIRE{Minimal sample size $n$, minimal solver $\textit{PnP}$, matches $\mathcal{M}=\{\mathcal{U}\cup \mathcal{W}\}$,
max. number $F$ of sample trials, max. number $T$ of  iterations, inlier threshold $\sigma$}%
\REPEAT \label{line:repeat}
    \STATE $\mathcal{S}=\emptyset$, $f=0$
    \STATE Randomly sample $m_1 = s_1 \leftrightarrow p_1$ from $\mathcal{U}$
    \STATE $\mathcal{S}\leftarrow m_1$
    \WHILE{$|\mathcal{S}| < n$ \AND $f<F$} 
            \STATE Randomly sample  $m_i = s_i \leftrightarrow p_i$ from $\mathcal{U}$
            \IF{$\mathcal{C}(p_i)\cap \mathcal{C}(p_1) \neq \emptyset$}
            \STATE $\mathcal{S}\leftarrow m_i$
            \ELSE
            \STATE $f \leftarrow f+1$
            \ENDIF
    \ENDWHILE
    \IF{$|\mathcal{S}|<n$} 
    \STATE Jump to next iteration at line \ref{line:repeat}
    \ENDIF
    \STATE Compute pose $\theta = \textit{PnP}(\mathcal{S})$
	\FORALL{$m_i\in \mathcal{M}$}
		\IF{$e(m_i;\theta)<\sigma$}
		\STATE $Inliers(\theta)\leftarrow Inliers(\theta)+1$
		\ENDIF
	\ENDFOR
	\IF{$Inliers(\theta)>Inliers(\theta^*)$}
		\STATE $\theta^* \gets \theta$
	\ENDIF
\UNTIL{$T$ iterations are reached}
\RETURN{best model $\theta^*$}
\end{algorithmic}
\end{algorithm}

\section{Experimental Evaluation}
\label{sec:experiments}
In this paper, we focus on a hybrid scene compression that enables usage of a pre-computed 3D sparse scene for visual localization.
Thus, especially for computationally constrained platforms, we are interested in: \begin{inparaenum}\item the run-time of the compression procedure, \item the memory reduction with respect to using an uncompressed scene, \item the localization rate, and \item the accuracy of the obtained poses\end{inparaenum}.

In order to properly validate the proposed method, we evaluate it with 6 different real-world datasets (\cf Tab.~\ref{table:datasets}).
We chose these datasets since they are representative of small- to medium-scale scenes and the use-cases we aim to tackle with our method.
By including datasets that have a wide variation in the number of points and imaging conditions, we show that our methods can work well in different scenarios.
For the evaluation, we use SIFT~\cite{lowe2004distinctive} descriptors throughout and use a single 6K-word visual vocabulary, pre-computed using the Dubrovnik dataset~\cite{li2010location}.
In practice, we did not observe strong evidence that suggested that using a vocabulary trained per dataset yielded consistently better results. This is also shown in \cite{sattler2017efficient}.
For the pose computation, we use the P4P solver by \cite{BujnakKP08}.

This section is organized as follows: first, we do an ablation study that analyzes the impact of the compression method introduced in \secref{section:compression} and the RANSAC variant of \secref{section:ransac}. Next, the best setting of our method is compared against previous works,  in terms of compression run-time, localization  rates, and localization accuracy. 
\begin{table}[t]
\centering
\scriptsize{
\begin{tabular}{|c|c|c|c|c|c|c|}
\hline
\multirow{2}{*}{Grid} & \multirow{2}{*}{MR} & \multicolumn{2}{c|}{Median time (ms)} & \multirow{2}{*}{\begin{tabular}[c]{@{}c@{}}\% reg. \\ images\end{tabular}} & \multicolumn{2}{c|}{Median error} \\ \cline{3-4} \cline{6-7} 
  &  & Query & RANSAC &  & Pos. (m) & Rot (\degree) \\ \hline
  &   & 181 & 2.4 & 95.7\% & 2.09 & 0.42 \\ 
$\bullet$ &   & 182 & \textbf{1.5} & 96.0\% & 2.02 & 0.37 \\ 
  & $\bullet$ & 190 & 7.7 & 97.4\% & 2.09 & 0.39 \\ 
$\bullet$ & $\bullet$ & \textbf{191} & {6.5} & \textbf{97.6\%} & \textbf{1.97} & \textbf{0.35} \\ \hline
\end{tabular}%
}%
\vspace{4pt}
\caption{Comparison of all the variants of our method for the Dubrovnik dataset compressed to $1.5\%$ of its original size. Where a $\bullet$ in ``Grid'' means we use image subdivisions and a $\bullet$ in  MR stands for the usage of our modified RANSAC.}
\label{table:variations}
\end{table}

\subsection{Ablation Study}
Tab.~\ref{table:variations} compares four different settings of our method.
We test its performance with and without image subdivisions and with and without our modified RANSAC on the Dubrovnik dataset. 
When not using our modified RANSAC, we do not take into account $\mathcal{P''}$ and its corresponding multi-matches.
Tab.~\ref{table:variations} shows an improvement with each of the proposed modifications.
In the following, we refer to using both image subdivisions and our modified RANSAC as ``Ours''.

\subsection{Compression Run-time}
\label{subsec:experiments_runtime}
For each dataset in Tab.~\ref{table:datasets}, we compress the scene to 1.5\% of its size (factoring both the number of full-descriptor 3D points and compressed points into this memory budget) and compare the run-times of our method to the state-of-the-art in model compression \cite{cao2014minimal}.
We have chosen 1.5\% to be the default compression rate since it exhibits a good trade-off between performance and memory size. However, the relative speed-up in compression time of our method vs. \cite{cao2014minimal} does not vary for different compression rates.
Tab.~\ref{table:datasets} shows that we achieve consistently lower compression times than \cite{cao2014minimal}. This is due to the fact that, as explained in \secref{section:compression}, our method only needs to check the occupancy of a visual word to figure out the cost of selecting a 3D point. 
Our approach is thus more suitable in scenarios where the scene changes over time, \eg, due to seasonal changes~\cite{Sattler2018CVPR}, and the 3D model used for localization needs to be adapted frequently.

\begin{table}[t!]
\centering
\scriptsize{
\begin{tabular}{|c|c|c|c|c|c|}
\hline
\
\multirow{2}{*}{Dataset} & \multirow{2}{*}{\begin{tabular}[c]{@{}c@{}}\# DB\\ images\end{tabular}} & \multirow{2}{*}{\begin{tabular}[c]{@{}c@{}}\# 3D\\ points\end{tabular}} & \multirow{2}{*}{\begin{tabular}[c]{@{}c@{}}\# Query\\ images\end{tabular}} & \multicolumn{2}{c|}{\begin{tabular}[c]{@{}c@{}}Compression\\ Time [s]\end{tabular}} \\ \cline{5-6} 
 &  &  &  & \textbf{Ours} & KCD~\cite{cao2014minimal} \\ \hline
Dubrovnik  \cite{li2010location} & 6,044 & 1.88M & 800 & \textbf{27.9} & 50.4 \\
Aachen  \cite{sattler2012image} & 3,047 & 1.54M & 369 & \textbf{21.6} & 45.5 \\
King's College  \cite{kendall2015posenet} & 1,220 & 503K & 343 & \textbf{1.7} & 2.5 \\
Old Hospital  \cite{kendall2015posenet} & 895 & 308K & 182 & \textbf{3.6} & 12.5 \\
Shop Facade  \cite{kendall2015posenet} & 231 & 82K & 103 & \textbf{0.43} & 1.15 \\
St Mary's Church  \cite{kendall2015posenet} & 1,487 & 667K & 530 & \textbf{8.74} & 26.1 \\ \hline
\end{tabular}
}
\caption{Datasets used and compression times (to compress the model to 1.5\% of the original points) for our method compared to ``KCD'' \cite{cao2014minimal} (using their implementation). The run-times for our method are significantly faster than those of \cite{cao2014minimal}.}
\label{table:datasets}
\end{table}

\begin{figure*}[t]
\includegraphics[width=\linewidth]{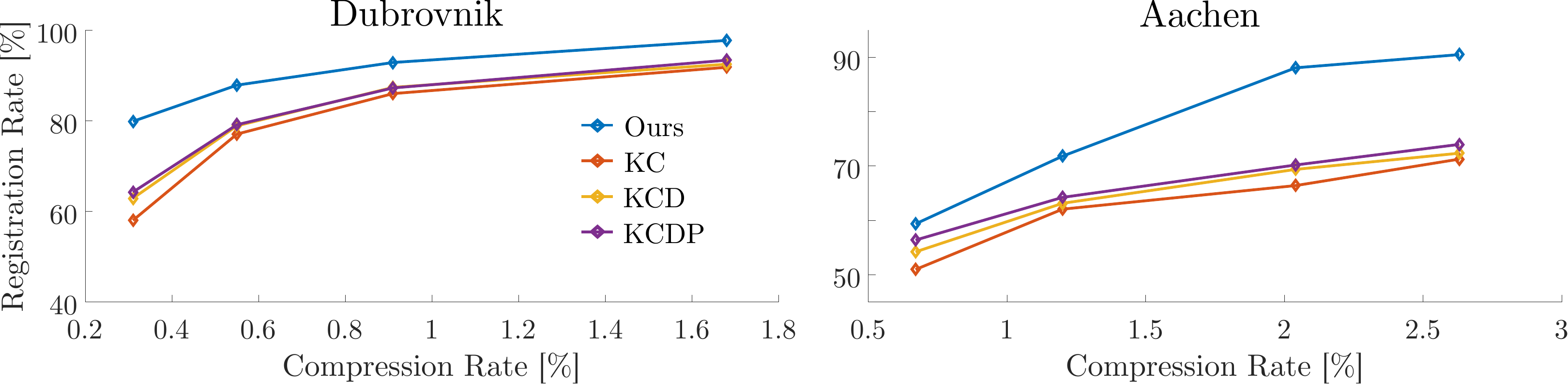}
\caption{Comparison of the registration rates (percentage of localized query images) for different compression methods at different compression rates. Our method consistently outperforms the current state of the art approaches from~\cite{cao2014minimal} for all compression rates.}
\label{fig:reg_rates}
\end{figure*}

\subsection{Registration Rates}
For the second experiment (see Fig.~\ref{fig:reg_rates}), we want to evaluate the efficiency of our compression and pose estimation methods in terms of image registration power.
Following~\cite{cao2014minimal,li2010location,sattler2017efficient,li:eccv:2012}, we consider an image as registered if the best pose found by RANSAC has at least 12 inliers. 

To make the compression rate comparison with \cite{cao2014minimal} fair, we select a combination of compressed and non-compressed points in order to match the memory used by~\cite{cao2014minimal} in each experiment. This is done in the following way:
for a given budget of $M$ 3D points selected by \cite{cao2014minimal}, we divide the budget into two parts.
The first 75\% of the budget is utilized by selecting $0.75 M$ uncompressed points by performing our weighted $K$-cover.
The second 25\% of the budget is then spent on compressed 3D points for multi-matches.
The number of compressed points selected is actually larger than $0.25 M$, since we can store (on average) 26.5 compressed points for each full point (see \secref{section:compression}).
Thus, we select $0.25 \times 26.5 M$ compressed points.
For example, for a compression rate of 2.04\% for Aachen, \cite{cao2014minimal} selects 40,377 full 3D points while we select 30,282 full 3D points plus 267,517 compressed points for the same memory budget.
We experimented with different rates of compressed vs full 3D points, and found that a 1:3 ratio on average performed best over all datasets, although other splits might work better on individual datasets. 

As can be seen from Fig. \ref{fig:reg_rates}, our method vastly outperforms the state-of-the-art method by \cite{cao2014minimal}. 
This is due to our two contributions, namely using an additional large set of compressed points and a more even distribution of the selected points in image space, as these are the main differences between our method and \cite{cao2014minimal}.
Notice that all compared methods achieve consistently better performances for Dubrovnik than for Aachen.
This is due to the different acquisition modes of the two datasets.
For Dubrovnik, the database images come from an internet photo-collection (Flickr) and thus tend to cluster around touristic areas. Thus, good camera coverage can be achieved with fewer points. Query images for this dataset were obtained by randomly removing images from a larger 3D reconstruction. 
They thus follow a similar distribution as the database images and can be localized with relative ease.
For Aachen, the database images were taken more regularly to cover the area more completely, presenting less overlap between cameras. Thus, more points might be needed to properly cover all cameras. The query images also were taken separately and do not follow the same distribution as the database images, resulting in a harder localization scenario. 

\begin{table}[t]
\centering
\scriptsize{
\begin{tabular}{|c|c|c|c|}
\hline
Method & MB Used & \begin{tabular}[c]{@{}c@{}}\#reg.\\  images\end{tabular} & \begin{tabular}[c]{@{}c@{}}Median \\ pos. error\end{tabular} \\ \hline
\textbf{Ours (1.5\% 3D pts)} & \textbf{3.79} & {782} & {1.97m} \\
PoseNet \cite{kendall2017geometric} & $\sim$50 & 800 & 7.9m \\ 
DenseVLAD~\cite{Torii2015CVPR} & 94.44 & 800 & 3.9m \\
Camera Pose Voting \cite{zeisl2015camera} & 288.57 & 798 & 1.69m\\
Camera Pose Voting + RANSAC \cite{zeisl2015camera} & 288.57 & 794 & \textbf{0.56m}\\
City-Scale Localization~\cite{Svarm2017PAMI} & 288.57 & 798 & \textbf{0.56m} \\
Active Search \cite{sattler2017efficient} & 953  & 795  & 1.4m \\
\hline
\end{tabular}
}
\vspace{4pt}
\caption{Accuracy for the Dubrovnik \cite{Snavely2008} dataset. Our method achieves a comparable accuracy and registration rate at a significantly lower memory consumption.}
\label{table:acc_dubrovnik}
\end{table}

\begin{table*}[t]
\centering
\scriptsize{%
\begin{tabular}{|c|ccc|ccc|ccc|ccc|}
\hline
\multirow{2}{*}{\textbf{Method}} & \multicolumn{3}{c|}{\textbf{King's College \cite{kendall2015posenet}}} & \multicolumn{3}{c|}{\textbf{Old Hospital \cite{kendall2015posenet}}} & \multicolumn{3}{c|}{\textbf{Shop Facade \cite{kendall2015posenet}}} & \multicolumn{3}{c|}{\textbf{St Mary's Chruch \cite{kendall2015posenet}}} \\ \cline{2-13} 
 & \multicolumn{1}{c|}{\begin{tabular}[c]{@{}c@{}}MB\\ used\end{tabular}} & \multicolumn{1}{c|}{\begin{tabular}[c]{@{}c@{}}\# reg.\\ images\end{tabular}} & \begin{tabular}[c]{@{}c@{}}Median \\ errors {[}m,\degree{]}\end{tabular} & \multicolumn{1}{c|}{\begin{tabular}[c]{@{}c@{}}MB\\ used\end{tabular}} & \multicolumn{1}{c|}{\begin{tabular}[c]{@{}c@{}}\# reg.\\ images\end{tabular}} & \begin{tabular}[c]{@{}c@{}}Median\\ errors {[}m,\degree{]}\end{tabular} & \multicolumn{1}{c|}{\begin{tabular}[c]{@{}c@{}}MB\\ used\end{tabular}} & \multicolumn{1}{c|}{\begin{tabular}[c]{@{}c@{}}\# reg.\\ images\end{tabular}} & \begin{tabular}[c]{@{}c@{}}Median\\ errors {[}m,\degree{]}\end{tabular} & \multicolumn{1}{c|}{\begin{tabular}[c]{@{}c@{}}MB\\ used\end{tabular}} & \multicolumn{1}{c|}{\begin{tabular}[c]{@{}c@{}}\# reg.\\ images\end{tabular}} & \begin{tabular}[c]{@{}c@{}}Median\\ errors {[}m,\degree{]}\end{tabular} \\ \hline
\textbf{Ours (@ 1.5\%)} & \textbf{1.01} & {343} & {0.81, 0.59} & {0.62} & {178} & {0.75, 1.01} & {0.16} & {103} & {0.19, 0.54} & {1.34} & {530} & {0.50, 0.49} \\
DenseVLAD~\cite{Torii2015CVPR} & 10.06 & 343 & 2.80, 5.72 & 13.98 & 182 & 4.01, 7.13 & 3.61 & 103 & 1.11, 7.61 & 23.23 & 530 & 2.31, 8.00\\
PoseNet \cite{kendall2015posenet} & 50 & 343 & 1.92, 5.4 & 50 & 182 & 2.31, 5.38 & 50 & 103 & 1.46, 8.08 & 50 & 530 & 2.65, 8.48 \\
Bayes PoseNet \cite{kendall2016modelling} & 50 & 343 & 1.74, 4.06 & 50 & 182 & 2.57, 5.14 & 50 & 103 & 1.25, 7.54 & 50 & 530 & 2.11, 8.38 \\
LSTM PoseNet \cite{walch2017image} & $\sim$50 & 343 & 0.99, 3.65 & $\sim$50 & 182 & 1.51, 4.29 & $\sim$50 & 103 & 1.18, 7.44 & $\sim$50 & 530 & 1.52, 6.68 \\
$\sigma^2$ PoseNet \cite{kendall2017geometric} & $\sim$50 & 343 & 0.99, 1.06 & $\sim$50 & 182 & 2.17, 2.94 & $\sim$50 & 103 & 1.05, 3.97 & $\sim$50 & 530 & 1.49, 3.43 \\ 
geom. PoseNet \cite{kendall2017geometric} & $\sim$50 & 343 & 0.88, 1.04 & $\sim$50 & 182 & 3.20, 3.29 & $\sim$50 & 103 & 0.88, 3.78 & $\sim$50 & 530 & 1.57, 3.32 \\ 
DSAC++ \cite{Brachmann2018CVPR} & ~207 & - & \textbf{0.18, 0.3} & ~207 & - & \textbf{0.20, 0.3} & ~207 & - & \textbf{0.06, 0.3} & ~207 & - & \textbf{0.13, 0.4} \\
Active Search \cite{sattler2017efficient} & 275 & 343 & 0.57, 0.7 & 140 & 180 & 0.52, 1.12 & 38.7 & 103 & 0.12, 0.41 & 359 & 530 & 0.22, 0.62 \\ 
\hline
\end{tabular}
}%
\vspace{4pt}
\caption{Comparison with several state-of-the-art methods on the Cambridge Landmarks datasets~\cite{kendall2015posenet}.}
\label{table:acc_cambridge}
\end{table*}

\subsection{Localization Accuracy}
\label{experiments:accuracy}
Finally, we want to focus on the impact our scene compression method has on the accuracy of the resulting camera poses.
To this end, we compare to state-of-the-art feature-based methods \cite{sattler2017efficient,zeisl2015camera,Svarm2017PAMI}, recent deep-learning-based methods \cite{kendall2015posenet,kendall2016modelling,kendall2017geometric,walch2017image,Brachmann2018CVPR}, and an image retrieval approach using compact image-level descriptors~\cite{Torii2015CVPR}. 
Results are shown for the Dubrovnik \cite{Snavely2008} dataset in Tab.~\ref{table:acc_dubrovnik} and the Cambridge Landmarks datasets~\cite{kendall2015posenet} in Tab.~\ref{table:acc_cambridge}.
The methods in \cite{sattler2017efficient,zeisl2015camera,li2010location,Svarm2017PAMI} are similar to ours in that they also make use of SIFT features to localize a given query image to a sparse 3D scene. For these approaches, we are able to accurately compute the memory consumption for representing the scene, but ignore potential overhead due to search structures.
For the deep learning methods, an approximate size of the network is provided.

It should be noted that the deep learning methods do not intrinsically provide a way to judge if a particular queried image was successfully localized (whereas in geometric methods one can rely on, \eg, inlier statistics). Thus, registration rates cannot be fairly compared to our approach. 
The registration rates for DSAC++ are not reported in~\cite{Brachmann2018CVPR}. 

As can be seen in Tab.~\ref{table:acc_dubrovnik} and Tab.~\ref{table:acc_cambridge}, our method clearly presents the best trade-off between memory consumption and performance. 
Compared to the retrieval baseline~\cite{Torii2015CVPR} and the pose regression techniques~\cite{kendall2015posenet,kendall2016modelling,walch2017image,kendall2017geometric}, our approach achieves both a lower memory consumption and higher pose accuracy. Compared to Active Search~\cite{sattler2017efficient}, our approach provides a comparable pose accuracy while achieving a reduction in memory consumption by more than order of magnitude. DSAC++~\cite{Brachmann2018CVPR} estimates poses that are significantly more accurate than those computed by our approach. However, DSAC++  is not able to provide a compact representation for small scenes. 
Note that the high pose accuracy obtained by~\cite{zeisl2015camera} and~\cite{Svarm2017PAMI} on the Dubrovnik dataset comes at a high run-time costs of more than 2 seconds per image on average.

Overall, our results show that our method represent the best of both worlds: it has a very small memory footprint (up to orders of magnitude lower than in the non-compressed setting) while achieving a performance close to the state-of-the-art feature-based methods for efficient localization.

\section{Conclusion}
In this paper, we have proposed a hybrid 3D scene compression scheme and a RANSAC variant that jointly manage to efficiently and accurately localize a given query image, all while using a small fraction ($\sim1.5\%$) of the original memory requirements.
Differently to previously proposed methods, we compress the 3D scene by producing two disjoint sets of 3D points: 
1) a set of points with their full visual descriptors that can be used to produce one-to-one matches given a query descriptor and 
2) a second set of points for which we only store a compressed descriptor. This second set of points is used to produce one-to-many matches, \ie, multi-matches. 
The hybrid output of our compression scheme is carefully selected to ensure a good scene coverage and visual uniqueness of the selected 3D points.
To properly handle the special two-fold output of our compression, we design a RANSAC variant to handle multi-matches such that they are only used during the model verification step, thus increasing number of inliers without negatively affecting the number of iterations. 
In addition, we propose a highly efficient co-visibility-based strategy to guide sampling within RANSAC. 

We have validated our approach using several real-world datasets. 
Our results show that our method achieves localization rates and a pose accuracy comparable to state-of-the-art feature-based and CNN-based methods at a significantly lower memory footprint. 

{
\PAR{Acknowledgements.} We thank Google's Visual Positioning System for their support.
}

{\small
\bibliographystyle{ieee_fullname}
\bibliography{IEEEabrv}
}

\end{document}